# Entity Matching by Pool-based Active Learning

Youfang Han[1], Chunping Li[1*]

[1] School of Software, Tsinghua University, Beijing, China

**\* Correspondence:**

Chunping Li, cli@tsinghua.edu.cn

## Abstract

The goal of entity matching is to find the corresponding records representing the same real-world entity from different data sources. At present, in the mainstream methods, rule-based entity matching methods need tremendous domain knowledge. The machine-learning based or deep-learning based entity matching methods need a large number of labeled samples to build the model, which is difficult to achieve in some applications. In addition, learning-based methods are easy to over-fitting, so the quality requirements of training samples are very high. In this paper, we present an active learning method ALMatcher for the entity matching tasks. This method needs to manually label only a small number of valuable samples, and use these samples to build a model with high quality. This paper proposes a hybrid uncertainty as query strategy to find those valuable samples for labeling, which can minimize the number of labeled training samples meanwhile meet the task requirements. The proposed method has been validated on seven data sets in different fields. The experiment shows that ALMatcher uses only a small number of labeled samples and achieves better results compared to existing approaches.

## 1. Introduction

Entity matching aims to determine whether the corresponding data records in different data sources represent the same entity. For example, there are two data sets of resident information in Figure 1, the goal is to judge whether the two records represent the same resident from two different data sets by comparing the attributes (name and age). Entity matching has a broad range of applications, and there is a lot of work devoted to the development of the entity matching system [1,2]. For instance, Mellgan [3] is the most advanced open-source entity matching solution (EM solution) in recent years. It constructs a complete entity matching system, which can be directly used by users for data cleaning and data integration.

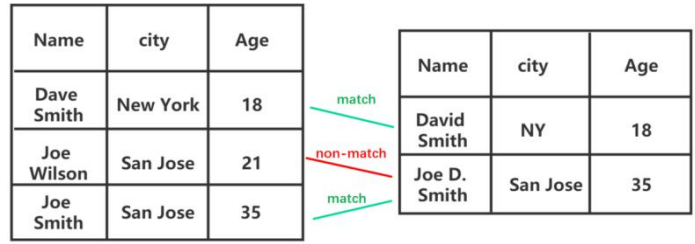

Figure 1: An example of entity matching

At present, mainstream entity matching methods are mostly rule-based or learning-based [4]. Rule-based entity matching methods usually require users to have a certain understanding of the data set, or need experts in the related domain to design the rules for better matching effects. These methods are sensitive to the selection of similarity measure method and threshold, which usually needs special design by domain experts [5,6,7]. Learning-based entity matching methods include machine-learning based (ML-based) and deep-learning based (DL-based) entity matching. The ML-based methods use the existing powerful machine learning algorithms to automatically learn the characteristics of matched entities [8,9]. DL-based methods [10,11] usually need pre-trained language models (LM) and domain-related texts for fine-tuning. Although machine learning and deep learning techniques have achieved good performance in entity matching tasks, they still have some limitations in practical application. In many actual scenarios, it is difficult to obtain a large number of labeled samples. Manual labeling requires a lot of labor-cost and time-cost, and usually difficult to obtain adequate effective labels in a short time quickly. Entity matching tasks usually have extremely imbalanced data samples. Generally, the number of mismatched samples is much larger than that of matched samples. The binary classification with imbalanced label distribution may lead to insufficient training of matched samples.

For entity matching, the benefit of labeling a large number of entity pairs is low. For example, "journey to the west" and "pilgrimage to the west" represent the same entity, "dream of the Red Mansions" and "story of the stone" also represent the same entity. The former pair can be judged to be the same entity through simple character-level comparison, while the latter pair can be judged only through labeling by relevant experts. This phenomenon is very common for entity matching. Many entity pairs can determine them matched or mismatched by easy comparison, so the benefit of labeling these data is very low. Therefore, if the records in the two data sets are directly handed over to experts for pairwise labeling, there will be a lot of extraordinary workloads. Although the deep learning method based on LM can achieve a good matching result, it usually needs suitable pre-trained LM and domain-related text to fine-tuning. When encountering new domain problems, this method is difficult to achieve good results without pre-trained LM and domain-related data.

This paper proposes an active learning method training the entity matching model to solve the problems as mentioned above. The aim of the proposed method is to label the least number but most meaningful samples, and achieve high accuracy at the same time. An active learning system attempts to label some unlabeled entities by asking Oracle (such as manual annotators), to overcome the bottleneck of scarce labels [12]. An effective active learning algorithm can achieve exponential acceleration of labeling efficiency [13,14]. Some application scenarios using this technology to solve the problem of lack of labeled samples or low-quality samples, such as information classification [15] and medical analysis [16], etc. The active learning method actively puts forward some labeling requests and submits some selected data to experts for labeling, which can not only greatly reduce the labeling-cost, but also effectively improve the quality of labeled samples, reduce the impact of noise data, and improve the generalization ability of learning model.

In this paper, we propose a pool-based active learning method integrating with query strategies, named as ALMatcher. ALMatcher can efficiently deal with the entity matching tasks on a few labeled training samples without domain knowledge. We design a hybrid uncertainty as query strategy, which can select the most valuable samples from the unlabeled data pool for labeling. ALMatcher does not need pre-trained LM and complicated pre-processing as well, and has good performance with a few labeled training samples by traditional classifiers.

The contributions of the paper are summarized as follows:

1. We propose a pool-based active learning method ALMatcher applied to entity matching. ALMatcher can find the most valuable labeled samples to build the learning model using only a small number of labeled samples and achieve good performance. To our best knowledge, this is the first work based on the active learning method for the entity matching tasks, which can effectively solve the problems in the acquisition of labeled samples and lack of domain knowledge in the application scenarios of entity matching.

2. Our method integrates with query strategies to select the valuable samples effectively from unlabeled samples for labeling. Experiments show that the selected samples are highly representative and the method can effectively reduce the labeling workload.

3. We validate the performance of our method on seven public datasets. Compared to current existing methods, ALMatcher can reach a similar F1 score, meanwhile use only small number of labeled data. In some scenarios of small-scaled data sets, ALMatcher is even superior to the state-of-the-art deep learning methods.

## 2. Related Works

This section introduces the related works of entity matching tasks, including rule-based entity matching, machine learning based entity matching, and deep learning based entity matching.

### 2.1 Rule based Entity Matching

Rule-based entity matching methods usually need a lot of domain knowledge to design. With the support of domain experts, the rule-based methods can perform well on most entity matching tasks. For example, on secure data sharing [17], mobile edge search [18], and other fields, the rule-based entity matching methods can well meet the task requirements. Besides, [19] proposed a powerful tool to automatically generate rules that satisfy a given high level specification, which make rules easy to design.

When we cannot get complicated rules designed by experts, the method of Linearly Weighted Combination Rules (LWCR) [4] can be used. LWCR is a useful baseline method for entity matching, and our proposed method will use LWCR as an auxiliary for measuring the similarity of entity pairs. LWCR is to weight the sum of the similarity values of each attribute, as shown in Formula 1.

$$\text{Sim}(x, y) = \sum_{i=1}^{n} \alpha_i \cdot s_i(x, y) \tag{1}$$

where x and y represent records from two different data sets, and n represents the number of attributes. $\alpha_i$ represents the preset weight, i.e., the importance of the i-th attribute. The sum of $\alpha_1$ to $\alpha_n$ is 1. $s_i(x, y)$ represents the similarity of the i-th attribute of x and y, which value is between 0 to 1. $s_i(x, y)$ usually adopts existing string-similarity measure methods, such as Levenshtein method [20], Jaro-Winkler method [21], and Jaccard method [22], etc.

The entity matching method based on LWCR has some limitations. First, it is difficult to set the weight of each attribute, which needs the user to know the importance of each attribute. Second, the relationship between attributes and similarity may not linear. However, LWCR can get an approximate similarity of each record pair quickly without much domain knowledge. Although the similarity calculated by LWCR may not be accurate, it can provide an approximate result to judge whether the record pair is matched. This approximate result will be used as the standard for pruning the training set and the standard for selecting the initial labeled data pool in the subsequent experiments.

## 2.2 Machine Learning based Entity Matching

The machine learning based (ML based) method can be used to predict whether the record pair is matched. We can train a model M for prediction through the train set, and then apply M to predict on the test set to judge whether the record pair is matched. Let the train set T = {($x_1$, $y_1$, $l_1$),..., ($x_n$, $y_n$, $l_n$)}. $x_i$ and $y_i$ represent a record pair, $l_i$ represents the label, in which "yes" is matched and "no" is mismatched.

To facilitate the training of the ML model, it is necessary to pre-process the record pair ($x_i$, $y_i$). Define a set of features f to quantify the record pairs, and then T can be transformed into T' = {(<$f_1(x_1,y_1)$,..., $f_m(x_1,y_1)$>, $c_1$), ..., (<$f_1(x_n,y_n)$,..., $f_m(x_n,y_n)$>, $c_n$)}. m indicates the number of features extracted, n indicates the number of groups of record pairs. c indicates the label, i.e. $c_i$ = 1 indicates $l_i$ = "yes", $c_i$ = 0 indicates $l_i$ = "no". The feature f is usually defined by the similarity of each attribute. After obtaining the transformed data set T', we can apply existing ML models for training, such as SVM, random forest, and so on.

At present, a series of works are based on ML models for entity matching tasks [23,24,25,26]. [25] compares the performance of different models on entity matching tasks. [26] proposes a stacking approach for threshold-based ML models, i.e. using integrated method to improve the prediction effect. The ML based methods are suitable for most application scenarios, but it depends on the data quality. When the unbalanced sample distribution, noise samples and fewer labeled samples occur in data sets, the effect of ML methods will be greatly degraded

### 2.3 Deep Learning based Entity Matching

With the development of deep learning technology, the deep learning based (DL based) entity matching methods gradually show their advantages. DeepER [27] and DeepMatcher [28] are the state-of-the-art DL based methods that have achieved good performance on entity matching tasks. DeepER uses LSTM-based RNN with the Siamese architecture. DeepER is the first method that tokenizes each record pair using embedding technology, such as Glove [29] and fastText [30], and then aggregates token-level embeddings into an entity representation. DeepMatcher also uses RNN to build a hybrid Sequence-aware with Attention model. Unlike DeepER, DeepMatcher calculates the similarity of attributes from the input record pairs to capture the similarity at the attribute-level. Meanwhile, [28] compared the results of different neural networks, such as SIF [31], RNN [32], and Attention [33]. These DL based methods have achieved good performance in entity matching tasks. Although the DL based methods can get better prediction result than traditional machine learning methods, they need a lot of labeled data to support. For small-scale, few-labels data sets, the DL based methods are difficult to achieve their best performance.

Besides, some innovative methods use pre-trained language models (LMs) and domain-related text on entity matching tasks. [34,36] proposed a novel entity matching system Ditto, based on pre-trained Transformer-based LM. This method fine-tunes pre-trained LM Sentence-BERT [35] for entity matching tasks, which allows domain knowledge to be added by highlighting important pieces of the input that may be useful for matching decisions. [37] and [38] also applied pre-trained language models to entity matching tasks, which showed that a pre-trained LM can further boost performance. In addition to pre-trained LM, [39] adopts transfer learning on entity matching tasks. This method uses pre-trained entity matching models from large-scale, production knowledge bases (KB), and then fine-tunes a small number of samples to get a good prediction performance. They all need the assistance of pre-trained LM and domain-related knowledge. When the application scenarios without domain knowledge or the pre-trained LM is not suitable, it is hard to get the best effect of these methods.

## 3. The Pool-based Active Learning Method for Entity Matching

This section introduces ALMatcher, i.e. the proposed active learning method for entity matching. The purpose of ALMatcher is to use the least number of labeled record pairs for prediction and get good results meeting requirements. We will introduce the overall process of the algorithm in subsection 3.1, the pre-process method in subsection 3.2, the method of generating the initial labeled pool in subsection 3.3, the query strategies in subsection 3.4, and the stop criterion in subsection 3.5.

Table 1: Essential Symbols

| Symbol | Definition |
| --- | --- |
| M | Iteration Number |
| S | Set of labeled record pairs on train set ( Labeled data pool) |
| Q | Set of all record pairs on train set |
| $Q_n=\{q_1, q_2, ..., q_n\}$ | Set of n queried record pairs |
| $L(q_1), L(q_2), ..., L(q_n)$ | Labels provided by the expert |
| $C=\{C_1, C_2, ..., C_n\}$ | Set of classifiers |
| T | Set of labeled record pairs on the test set |
| V | Set of labeled record pairs on valid set |
| $F1(V, C_i)$ | F1 score w.r.t. ground truth of a set V and classifier $C_i$ predicts labels for elements in V |
| $C_{i,j}$ | Classifier $C_i$ in iteration j |

Some essential symbols are defined in Table 1. S represents the labeled data pool, i.e. the set of labeled record pairs, which are used to construct classifiers, and its size is | S |. M represents the maximum number of iterations, i.e. M labeling queries. Q represents all record pairs in the train set, and | Q | is its size. $Q_n$ indicates that a group of record pairs $\{q_1, q_2,..., q_n\}$ is labeled in an iteration. $L(q_1), L(q_2), ..., L(q_n)$ indicates that the labels of the record pairs in $Q_n$. Formula L: Q → {0,1} is regarded as the ground truth, i.e. the result labeled by experts, where 0 indicates mismatched and 1 indicates matched. V represents all record pairs in the validation set. The record pairs in V have been labeled. The data in V is used to evaluate the performance of the intermediate models in the process of active learning. T represents all record pairs in the test set, and the data in T is used for the evaluation of the final model. F1 (V, $C_i$) represents the F1 score calculated by the labeled data in V using the classifier $C_i$. F1 score is used to measure the performance of classifier $C_i$ on the test set and validation set. After m iterations, the prediction result of the best classifier can be expressed as F1 (T, $C_i$, m).

## 3.1 The Framework of ALMatcher

The active learning method selects the most valuable samples from the unlabeled samples continuously and submits them to the experts for labeling, and then verifies whether the model performance meets the requirements after adding new labeled data. If it does not meet the requirements, continue to select the most valuable samples and submit them to the experts for labeling. The value of samples is usually judged with uncertainty. If the uncertainty of a sample is higher, it means that the sample needs to be judged by experts manually, and these samples have a higher probability to improve the performance of the model. The sample query strategies of active learning are broadly divided into two categories: stream-based strategies [40] and pool-based strategies [41]. For stream-based strategies, unlabeled samples will be handed over to the selection engine in order, and the selection engine will decide whether to submit the sample for labeling. The sample will be discarded if it not be selected. The pool-based strategies will maintain a set of labeled data pool, and each time the selection engine will select the most valuable samples from the unlabeled data to labeled data pool.

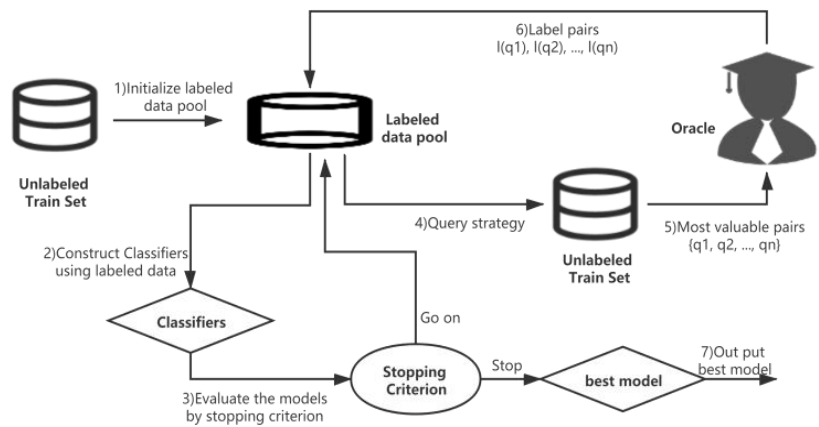

Figure 2：The work process of ALMatcher on entity matching

ALMatcher for entity matching adopts the pool-based query strategy, which process is shown in Figure 2. Firstly, select a small number of valuable samples actively from unlabeled training samples for labeling, i.e. construct the initial labeled pool. Then, construct the initial classifiers by labeled samples, and the performance of the classifiers is evaluated by the stop criterion. If the performance of classifiers meets the requirements, output the optimal classifier, otherwise, calculate the uncertainty of each sample from the unlabeled train set, select n samples with the most uncertainty for labeling according to the query strategies, and add these labeled samples to the labeled pool. Repeat the above steps until the classifier meets requirements, i.e. stop criterion. In each iteration, the label-request will be put forward actively, and the most valuable samples will be selected and handed over to the experts for labeling. This process can maximize the benefit of the experts' work.

The active learning algorithm for entity matching is described in Figure 3. The part of the data preprocessing includes structuring the data set and pruning the train set (lines 1-2). In the part of active learning, generate the initial labeled pool firstly (line 3), and then start the iteration of active learning. In each iteration, the first step is building the ML models with the data in the labeled pool, and then using these intermediate models to predict the data in the validation set. When the stop criterion is met, exit the iteration (lines 4-7), otherwise, the most valuable samples from the unlabeled train set are selected for labeling according to the query strategies, and these labeled samples are added to the labeled pool (line 8-9). When the iteration is terminated, output the optimal model (line10).

```
1. Generate data set
2. Pruning
3. Generate Initial labeled pool S from train data Q
4. While True:
5.      Construct classifiers C using labeled data pool S
6.      if satisfy stop criterion:
7.          break
8.      Get Q_n={q_1, q_2, ..., q_n} by query strategy
9.      Add <q_1, L(q_1)>, <q_2, L(q_2)>, ..., <q_n, L(q_n)> to S
10. Return C_best
```

Figure 3. Algorithm of active learning for entity matching

## 3.2 Data Preprocessing

### 3.2.1 Constructing the feature vector

The original data set is usually unstructured. It is necessary to standardize the original data into structured data before active learning. As mentioned in subsection 2.2, each pair of records $(x_i, y_i)$ needs to be quantified with a set of features f to convert the data $(x_1,y_1,l_1)$ into $(<f_1(x, y),..., f_m(x, y)>, c)$. String-similarity measure methods mentioned in subsection 2.1 are used to construct feature f.

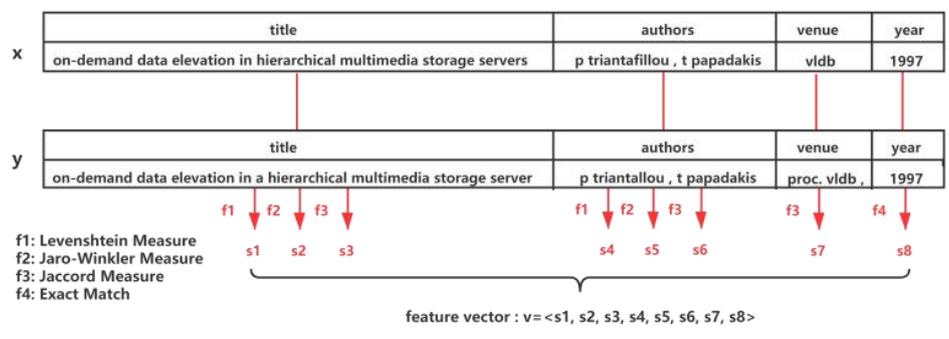

Figure 4: Construction of feature vector

Each attribute adopts one or more string-similarity measure methods to construct a feature, and all features are spliced together to form the feature vector $<f_1(x_1,y_1),..., f_m(x_1,y_1)>$. Different attributes need to be used in different string-similarity measure methods to construct the feature vector according to their characteristics. The standardization rules of the train set, validation set, and test set are the same. Taking Figure 4 as an example, the Levenshtein method, Jaro-Winkler method, and Jaccard method are used to calculate the similarity of attribute "title" and "author", Jaccard method is used to calculate the similarity of attribute "venue", and the exact matching method is used to calculate the similarity of attribute "year". Then, these similarity values are spliced into feature vectors v = $<s_1, s_2, ..., s_8>$ for model training.

### 3.2.2 Pruning

Entity matching is a binary classification problem, in which input is record pairs and output is "match" or "mismatch". Unlike most binary classification tasks, the sample distribution of entity matching is extremely imbalanced. Usually, a record in data source A can just match with one record in data source B. In large-scale data sets, every time a matched record pair is added, N + M mismatched record pairs will be added at the same time. N and M are the sizes of the two data sources respectively. In other words, the number of record pairs increases exponentially with the size of the data source, but the number of matched record pairs only increases linearly, which makes the number of negative samples far more than positive samples.

The imbalanced distribution of record pairs will lead to poor performance of matched samples prediction and over-fitting of mismatched samples when using model training directly. Prediction of matched samples is the key to entity matching, so it is necessary to reduce the number of mismatched samples on the train set as much as possible, meanwhile retain as many matched samples as possible. Blocking is an effective method to cut a large number of mismatched record pairs. At present, there are a series of works on blocking technology [42].

In subsection 2.1, the LWCR method can calculate an approximate similarity for record pairs. The accuracy of this method is low, but if the similarity of a record pair is very low and lower than a threshold, these record pairs have a high probability be mismatched. Therefore, the pruning strategy in our work is to calculate a similarity for all record pairs in the training set by LWCR, sort all record pairs according to the similarity value, and prune the pairs whose similarity is lower than a preset threshold. This pruning strategy can not only cut a large number of mismatched samples but also reduce the data noise. Some record pairs represent the same entity in the real world but have extremely low similarity. Although these record pairs are matched, their low similarity may lead to deviation in model training. The pruning strategy cuts such low similarity matched pairs at the same time, which can improve the generalization ability of learning models to some extent.

### 3.3 Generating the Initial Labeled Pool

At the beginning of active learning, we need to create an initial labeled pool, and the subsequent operations will maintain this labeled pool. The quality of initial labeled pool will directly affect the performance of initial classifiers and have a great impact on the whole process of active learning. If the performance of initial classifiers is very poor, it is difficult to select the valuable samples for labeling in the following iterations, which will lead to slow down the improvement of the model.

The simple method of generating the initial labeled pool is to select n record pairs randomly, but this method is unstable. In the case of an unbalanced distribution of samples, the probability of selecting mismatched pairs is higher, which leads to a poor effect of fitting. Meanwhile, this method may select many fuzzy samples, so that initial classifiers cannot learn the feature of matched samples and mismatched samples effectively.

Here we use the LWCR method to generate the initial labeled pool. First, calculate an approximate similarity of each record pair according to the LWCR, and then sort all record pairs according to this approximate similarity value, and take N / 2 samples from the head and the end respectively for constructing the initial labeled pool. The advantage of this method is that it can ensure the positive and negative samples of initial data are equal. Because the data with the highest similarity are matched, while the data with the lowest similarity are mismatched. Besides, the data at the head and at the end are the most representative, which can enable initial classifiers to learn basic features and make a preliminary judgment on the matched and mismatched record pairs effectively.

### 3.4 Query Strategy

The key of active learning is to select N samples with the highest value for labeling during each iteration. Samples with the highest value are those that have the greatest impact on model improvement. If the label of a record pair can be calculated easily, experts do not need to spend time labeling it, and adding it to the labeled pool cannot improve the model to the greatest extent. Conversely, if the intermediate model is difficult to judge the label of a record pair, e.g. the probability judges as matched is close to probability judges as mismatched, it is needed to get an accurate label by experts, and then add it to the labeled pool for a great improvement on classifiers.

The goal of the query strategy is to find the unlabeled record pairs with the highest uncertainty to improve the predictive model greatly. The classifiers in ALMatcher are probability models, which can calculate the probability of matched and mismatched respectively. The entropy is one of the most common methods for calculating uncertainty on the probability model [43]. This method only considers the prediction results of one classifier. The entropy uncertainty calculated from one classifier is usually inaccurate. Therefore, it is necessary to consider the prediction results of different classifiers comprehensively for more accurate uncertainty. We further propose a hybrid uncertainty as query strategy, which integrates three parts, i.e., entropy-average uncertainty, entropy-variance uncertainty, and probability-variance uncertainty.

#### 3.4.1 Entropy-average uncertainty

Each iteration of active learning will generate multiple classifiers to predict each record pair, and each classifier will generate the probability of "matched" and "mismatched" and calculate the entropy of the samples. $H_1(e)$, $H_2(e)$, ..., $H_n(e)$ represent the information entropy of n classifiers $C_1$, $C_2$, ..., $C_n$, respectively. The entropy-average uncertainty can consider the results of all classifiers comprehensively, which is defined in formula 2:

$$\mathbf{Ave\_Entropy(e)} = \frac{\sum_{i=1}^{n} \mathbf{H_i(e)}}{\mathbf{n}} \quad (2)$$

Intuitively, this method considers the entropy calculated by different classifiers comprehensively, i.e. the higher the average of entropy, the more difficult it is for the model to predict. The query strategy by entropy-average uncertainty selects n samples with the highest value of **Ave_Entropy(e)** for labeling.

### 3.4.2 Entropy-variance uncertainty

In some cases, using the entropy-average uncertainty as query strategy cannot select the most valuable samples. Some record pairs may have high entropy calculated in one classifier and low entropy calculated in another classifier. Comparing with the record pairs with moderate entropy calculated in both classifiers, the uncertainty of the former should be higher, even if the average entropy of the two record pairs is same. Therefore, we propose another method to calculate uncertainty, namely entropy-variance uncertainty, as shown in formula 3:

$$\mathbf{Var\_Entropy(e)} = \frac{\sum_{i=1}^{n}(H_i(e) - Ave\_Entropy(e))^2}{n} \quad (3)$$

This method can be used to evaluate the variance of entropy on different classifiers. If the entropy on one classifier is high and on the other classifier is low, the entropy variance of the record pair will be high and this record pair is more uncertain. This method can solve the limitation on entropy-average uncertainty.

### 3.4.3 Probability-variance uncertainty

There is still some limitations on entropy-variance uncertainty. Entropy can describe the uncertainty of samples. When one model predicts the sample as matched and another model predicts the sample as mismatched, they still may get the same entropy. These samples have low entropy-average uncertainty and entropy-variance uncertainty, but the uncertainty of them should be high, because the prediction results are totally different in these two models.

This phenomenon rarely occurs in most application scenarios. Because the result of record pairs predicted by different models are usually similar. However, the number of samples is relatively small in active learning. Adding samples with high uncertainty in each iteration may lead to deviation in the prediction of other samples by different models, which leads to some samples being predicted as different labels in different models. Since entity matching is a binary classification task, we can define the variance of the probability of matched samples predicted by different models as uncertainty, i.e., probability-variance uncertainty, as shown in formula 4:

$$\mathbf{Var\_Prob(e)} = \frac{\sum_{i=1}^{n}(p_{i,match} - \overline{p_{match}})^2}{n} \quad (4)$$

where $p_{i,match}$ represents the probability of classifier $C_i$ predicting sample e is matched, and $\overline{p_{match}}$ represents the average probability of all classifiers predicting that sample e is matched.

### 3.4.4 Hybrid uncertainty

The above-mentioned uncertainty calculation methods have their advantages and limitations as query strategies. We thus consider a hybrid query strategy that comprehensively combines the results of these three uncertainty calculation methods.

$$\mathbf{Hybrid(e)} = \sum \alpha_i Sort\_index(Uncertainty_i(e)) \quad (5)$$

First, calculate the entropy-average uncertainty, entropy-variance uncertainty, and probability-variance uncertainty of all unlabeled samples in Q, and sort these three uncertainties respectively. Then, the index sorted by these three methods is weighted sum up as hybrid uncertainty. Formula 5 shows the hybrid uncertainty, which $Uncertainty_i(e)$ means the methods of i-th uncertainty, α means the preset weight, and Sort_index means the index order of each uncertainty. The range of each index is from 1 to | Q |. If the weight of the three methods is 1, the lowest uncertainty of each instance is 3 and the highest uncertainty is 3 | Q |. The query strategy by hybrid uncertainty is selecting samples with the highest hybrid uncertainty. If the hybrid uncertainty of two samples is same, we compare the entropy-average uncertainty, i.e., selecting the samples with the highest entropy-average uncertainty. The weights of three uncertainty can be adjusted according to data characteristics. Generally, the weight of the entropy-average uncertainty is the highest. When there is a great dispute on the result of most samples by different models, the weight of entropy-variance uncertainty and probability-variance uncertainty can be increased appropriately. The hybrid uncertainty can evaluate the uncertainty of samples comprehensively. The subsequent experiments will compare the performance of hybrid uncertainty and the other three uncertainties.

### 3.5 Stop Criterion

In active learning tasks, an effective stop criterion is important. Stop criterion determines when the query process can be terminated. A good stop criterion should prevent the iteration from stopping prematurely while the performance of classifiers is still improved, and stop labeling samples when models are stable.

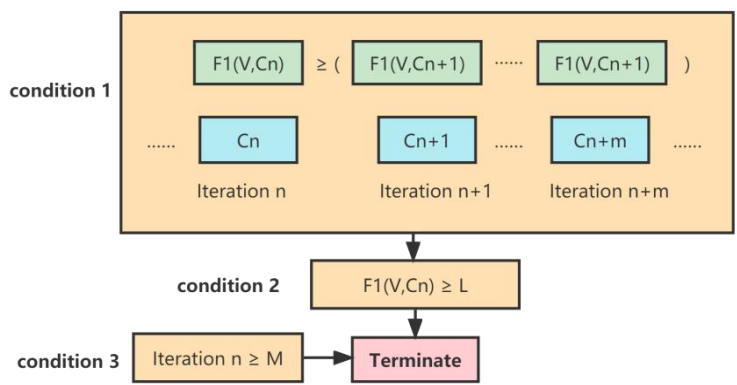

Figure 5 Stopping Criterion

Our stop criterion is shown in Figure 5. We use each intermediate classifier to predict the samples on validation set V. When the $F_1 (V, C_i)$ calculated by the best intermediate classifier C in the i-th iteration on validation set V does not increase or even decrease, the iteration will be terminated. Due to certain deviations in each intermediate classifier, we usually continue the active learning process for more stable classifiers, that is, if the F1 score does not increase after successive iterations, the classifier with the highest F1 score will be output. In addition, we need to set a minimum threshold L according to the result of other baseline methods. If the

F1 score predicted by the classifiers on the validation set V is lower than L, we will continue the iteration regardless of the stop criterion. This strategy can ensure that the result of active learning will not be too poor, and prevent the classifiers from stopping halfway before learning enough characteristics. Besides, the F1 score of intermediate classifiers may increase constantly when the labeled data are added continuously. We set an upper limitation M of iteration to avoid the iteration being terminated too late. When the number of iterations reaches M, the iteration will be terminated regardless of the stop criterion and lower bound L, and the best classifier will be output. If the size | S | of the labeled data pool is too large, it does not play the purpose of active learning, i.e. reducing the labeling workload. So, the upper limitation M of iteration is usually not too large.

## 4. Experiment and Evaluation

This Section will validate the effectiveness and applicability of the proposed ALMatcher through experiments. Subsection 4.1 introduces the experiment setup, Subsection 4.2 introduces the evaluation metrics for the experiment, and Subsection 4.3 shows the performance of ALMatcher through multiple experiments including the effectiveness of the query strategies, the method of generating the initial labeled pool, and the pruning method.

### 4.1 Experiment Setup

#### 4.1.1 Data Sets

Table 2 Information on experiment data sets

| Data Set | Domain | Pairs | match | # Attr. |
|---|---|---|---|---|
| Amazon-Google | software | 11460 | 1167 | 3 |
| BeerAdvo-RateBeer | Beer | 450 | 68 | 4 |
| DBLP-ACM | citation | 12363 | 2220 | 4 |
| DBLP-Scholar | citation | 28707 | 5347 | 4 |
| Fodors-Zagats | restaurant | 946 | 110 | 6 |
| iTunes-Amazon | music | 539 | 132 | 8 |
| Walmart-Amazon | electronics | 10242 | 962 | 5 |

The data used in our experiments are from seven public data sets for entity matching provided by [11]. The data sets are from different domains, as shown in Table 2. The column "pairs" lists the number of labeled samples in each data set. Each sample is a record pair composed of two records, labeled 1 for "matched" and 0 for "mismatched". The column "match" lists the number of matched samples. The column "# Attr." shows the number of attributes of the data set. All attributes are atomic, i.e., not a composite form of multiple values.

Table 3 Training set, testing set and validation set

| DataSet | Training Set | | Testing Set | | Validation Set | |
|---|---|---|---|---|---|---|
| | Match | pairs | match | pairs | match | pairs |
| Amazon-Google | 589/699 | 5077/6874 | 234 | 2293 | 234 | 2293 |
| BeerAdvo-RateBeer | 29/40 | 103/268 | 14 | 91 | 14 | 91 |
| DBLP-ACM | 1323/1332 | 2602/7417 | 444 | 2473 | 444 | 2473 |
| DBLP-Scholar | 2915/3207 | 6231/17223 | 1070 | 5742 | 1070 | 5742 |
| Fodors-Zagats | 62/66 | 227/567 | 22 | 190 | 22 | 189 |
| iTunes-Amazon | 74/78 | 166/321 | 27 | 109 | 27 | 109 |
| Walmart-Amazon | 536/576 | 5379/6144 | 193 | 2049 | 193 | 2049 |

These data sets have been divided into training set, validation set, and testing set according to the ratio of 6:2:2, and the labels have the same distribution. We use the training set for constructing models by active learning, use the validation set to verify the performance of intermediate classifiers, and finally use the testing set to verify the performance of the final classifier. Due to a large amount of data in the training set and the extreme imbalance distribution of matched samples and mismatched samples, we use the pruning strategy described in Subsection 3.2 to cut down most mismatched samples in the training set. The sample distribution of the pruned data sets is shown in Table 3. The left side of '/' in the train set represents the number of samples after pruning, and the right side of '/' represents the number of samples before pruning.

### 4.1.2 Classifiers

In ALMatcher, we use the mainstream machine learning algorithms to construct the classifiers, including SVM, random forest, KNN, and naive Bayesian, which are widely used in active learning tasks [12]. Record pairs are inputted into the classifiers and return "match" or "mismatch". The parameters and random seed of classifiers are consistent.

### 4.1.3 Hyper-parameters

In the experiment, the maximum number of iterations of active learning $M_{max}$ is 20. The seven data sets can be divided into small-scale data sets (BeerAdvo-RateBeer, iTunes-Amazon, and Fodors-Zagats ) and large-scale datasets (DBLP-ACM, DBLP-Scholar, Amazon-Google, Walmart-Amazon). For small-scale data sets, the size of the initial labeled pool | S | is 6, and each iteration adds 4 samples to the labeled pool. For large-scale data sets, the size of the initial labeled pool | S | is 50, and each iteration adds 20 samples to the labeled pool.

## 4.2 Evaluation Metrics

Due to the imbalanced distribution of samples on entity matching tasks, we use the F1 score [44] to measure the performance of models. F1 score makes a comprehensive judgment on the predicted results by calculating the harmonic average of precision and recall, as shown in Formula 6.

$$\mathbf{F1} = \frac{2*\mathbf{Precision}*\mathbf{Recall}}{\mathbf{Precision}+\mathbf{Recall}} \qquad (6)$$

In the experiment, we first compare the F1 score of different methods. The higher the F1 score, the better the performance of the method. When comparing the active learning methods of different query strategies, it is may occur two methods have the same F1 score. When the F1 score is same, we compare the size of labeled pool N. If N is small, it means that fewer labeled training samples are used, i.e. less labeling workload is required to achieve the same prediction effect.

### 4.3 Experiment Results

We validate the effectiveness of the query strategies and the strategy of initializing the labeled pool in the active learning stage, and the effectiveness of the pruning strategy on the training set in the preprocessing stage.

### 4.3.1 Performance of Query Strategies

We compare the performance of proposed query strategies and the entropy-based query strategy. In the experiment, we use the method of initializing the labeled pool proposed in subsection 3.3, and the stop criterion proposed in subsection 3.5. The integrated query strategies use four classifiers mentioned in subsection 4.1.2, and the entropy-based query strategy uses the random forest as the classifier. The parameters of the classifiers are consistent throughout the experiment.

Table 4 Comparing the performance of different uncertainty methods

| DataSet | Entropy | | Ave_Entropy | | Var_Entropy | | Var_Prob | | Hybrid | |
|---|---|---|---|---|---|---|---|---|---|---|
| | F1 | N | F1 | N | F1 | N | F1 | N | F1 | N |
| Amazon-Google | 0.337 | 450 | 0.348 | 430 | 0.376 | 450 | 0.209 | 450 | **0.424** | **370** |
| BeerAdvo-RateBeer | 0.786 | 38 | 0.800 | 38 | 0.800 | 50 | 0.759 | 62 | **0.815** | 62 |
| DBLP-ACM | 0.970 | 230 | 0.975 | 390 | 0.975 | 430 | 0.976 | 290 | **0.977** | 410 |
| DBLP-Scholar | 0.880 | 410 | 0.868 | 270 | **0.893** | **310** | 0.860 | 250 | 0.884 | 390 |
| Fodors-Zagats | 0.977 | 34 | **1.000** | **42** | 0.977 | 50 | 0.927 | 18 | **1.000** | 46 |
| iTunes-Amazon | 0.931 | 30 | **0.981** | **58** | 0.931 | 58 | 0.909 | 66 | 0.947 | 34 |
| Walmart-Amazon | 0.684 | 210 | 0.684 | 330 | 0.662 | 210 | 0.639 | 350 | **0.688** | 410 |

Table 4 compares the performance of proposed integrated query strategies and the baseline strategy based on entropy. The performance of the hybrid uncertainty is better than other uncertainties on four data sets, and the optimal F1 score is also obtained in dataset Fodors-zagats, in which the amount of labeled data is slightly higher than the entropy-average uncertainty. The entropy-average uncertainty performs best on two data sets, and the entropy-variance uncertainty performs best on one data set. The performance of probability-variance uncertainty is relatively poor than other integrated strategies. For most data sets, there are few noise data after pruning, so few samples lead to great deviation on intermediate classifiers. However, it improves the performance of hybrid uncertainty when adding probability-variance uncertainty as an optimization term.

Among these five query strategies, the hybrid uncertainty has the best performance overall. The entropy-average uncertainty and entropy-variance uncertainty can get good results on some data sets, but they are not as stable as the hybrid uncertainty. The performance of entropy-based uncertainty and probability-variance uncertainty is poor than the above three integrated strategies.

### 4.3.2 Compared to deep learning methods

To verify the effectiveness of ALMatcher, we compare ALMatcher with the mainstream ML based and DL based methods, i.e., SIF[31], RNN[32], Attention[33], DeepER[27], DeepMatcher[28], and Magellan[3]. The results of the optimal integrated query strategy corresponding to each data set in Table 4 are compared with these methods. The comparative results are shown in Table 5. Table 5 records the F1 score of each method in seven data sets. The last column "△F1" represents the difference in F1 score between ALMatcher and the optimal results of the other methods. Table 6 shows the number of labeled training samples for each method. The column "ratio" shows the ratio of labeled training samples used by ALMatcher to other methods.

Table 5 Comparing proposed method with other methods

| DataSet | SIF | RNN | Attention | Magellan | DeepER | DeepMatcher | ALMatcher | △F1 |
|---|---|---|---|---|---|---|---|---|
| Amazon-Google | 0.606 | 0.599 | 0.611 | 0.491 | 0.561 | **0.693** | 0.424 | -0.269 |
| BeerAdvo-RateBeer | 0.581 | 0.722 | 0.808 | 0.788 | 0.5 | 0.727 | **0.815** | 0.007 |
| DBLP-ACM | 0.975 | 0.983 | **0.984** | **0.984** | 0.976 | **0.984** | 0.977 | -0.007 |
| DBLP-Scholar | 0.909 | 0.93 | 0.933 | 0.923 | 0.908 | **0.947** | 0.893 | -0.054 |
| Fodors-Zagats | **1** | **1** | 0.821 | **1** | 0.977 | **1** | **1** | 0 |
| iTunes-Amazon | 0.814 | 0.885 | 0.808 | 0.912 | 0.88 | 0.88 | **0.981** | 0.069 |
| Walmart-Amazon | 0.651 | 0.676 | 0.5 | **0.719** | 0.506 | 0.669 | 0.688 | -0.031 |

From the comparing experiment results, the performance of ALMatcher is slightly lower than the DL-based method on datasets DBLP-ACM, DBLP-Scholar, and Walmart-Amazon, and the F1 score is lower than the optimal model of 0.03 on average. Meanwhile, the amount of labeled training samples of ALMatcher in these three data sets has decreased significantly, and the average amount of labeled training samples is only about 10% of other methods. In these three data sets, ALMatcher can greatly reduce the number of labeled samples while maintaining certain prediction accuracy. On datasets BeerAdvo-RateBeer, Fodors-Zagats, and iTunes-Amazon, ALMatcher is better than other methods. On these three datasets, our method can not only reduce the number of labeled training samples but also improve the F1 score.

Table 6: Comparing the number of labeled samples

| DataSet | ALMatcher | Other methods | Ratio |
|---|---|---|---|
| Amazon-Google | 370 | 5077 | 0.073 |
| BeerAdvo-RateBeer | 62 | 103 | 0.602 |
| DBLP-ACM | 410 | 2602 | 0.158 |
| DBLP-Scholar | 310 | 6231 | 0.050 |
| Fodors-Zagats | 42 | 277 | 0.152 |
| iTunes-Amazon | 58 | 166 | 0.349 |
| Walmart-Amazon | 410 | 5379 | 0.076 |

From Table 5 and Table 6, it can be found that ALMatcher performs better in small-scale data sets (the number of training samples is less than 1000), which can even get better results than the DL based methods. The possible reason is that the labeled data in small-scale data sets has a greater probability to represent the characteristics of the whole data set. In large-scale data sets (the number of training samples is greater than 1000), the amount of labeled training samples is too small comparing with the whole data set. So that a small amount of labeled data may not fit the characteristics of the whole data set, which makes the model perform worse than the DL-based methods. Nevertheless, ALMatcher is also applicable in large-scale data sets, which can greatly reduce the workload of labeling with losing little accuracy.

In the experiment, we also found that ALMatcher has poor performance on Amazon-Google. Amazon-Google records software-related data which has three attributes, i.e., "title", "manufacturer" and "price". Among these attributes, the missing rate of "manufacturer" is nearly 90%, and the "price" has no obvious matching law. Therefore, most of classifiers can only be constructed by the "title". In the case，only a single attribute is used to judge whether the record pair is matched or not, so it is difficult to obtain good results only by the method of string similarity, if the semantic information is not considered. ALMatcher only considers the structural similarity of the attribute, but for Amazon-Google, we need to analyze the semantic similarity to get more accurate prediction results. Therefore, the F1 score of Magellan and ALMatcher are far lower than DL-based methods.

### 4.3.3 Performance of Initial Labeled Pool

The quality of the initial labeled pool plays an important role in the active learning method. If the samples in the initial labeled pool are low-quality, it will lead to the poor effect of the first batch of classifiers, lead to the inaccurate entropy calculated by these classifiers, and the selected samples are not with the highest uncertainty.

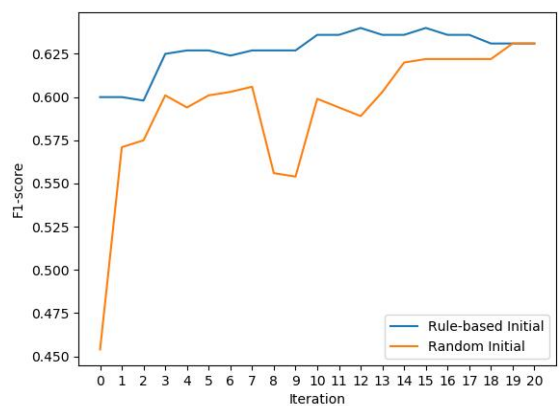

Figure 6 F1-Iteration curve on different initialization strategies

Figure 6 compares the performance of the random selection method and the rule-based method proposed in Section 3.3 for initializing the labeled pool on Walmart-Amazon. Without adding any new labeled samples, the F1 score of the rule-based method on the validation set has reached 0.6, while the random selection method is only 0.45. Using the same query strategy to add new labeled samples, the random selection method requires about 17 iterations to achieve the same F1 score as the rule-based method. There are similar results on other data sets. It can be seen that ALMatcher of initializing the labeled pool is effective, which can reduce the number of iterations and make the intermediate model fitting quickly.

### 4.3.4 Performance of Pruning Method

In subsection 2.1, we present a rule-based pruning method, i.e., using the linear weighted combination rule (LWCR) to calculate similarity for each record pair, and cutting the samples with low similarity. This method can remove a large number of mismatched pairs while retaining most of the matched pairs. The pruning effect on the train set can be seen in Table 2. The proportion of pruned matched pairs in all record pairs is only 3.52% on average, which shows that the pruning method is effective.

Table 7 Evaluate the performance of the pruning strategy

| DataSet | Hybrid | | Hybrid* | |
|---|---|---|---|---|
| | F1 | N | F1 | N |
| Amazon-Google | 0.282 | 450 | 0.424 | 370 |
| BeerAdvo-RateBeer | 0.759 | 26 | 0.815 | 62 |
| DBLP-ACM | 0.971 | 270 | 0.977 | 410 |
| DBLP-Scholar | 0.865 | 450 | 0.884 | 390 |
| Fodors-Zagats | 1.000 | 50 | 1.000 | 46 |
| iTunes-Amazon | 0.909 | 66 | 0.947 | 34 |
| Walmart-Amazon | 0.649 | 190 | 0.688 | 410 |

Table 7 compares the results of the proposed method using hybrid uncertainty as a query strategy before and after pruning. * indicates pruning operation. The results show that the F1 score of the method after pruning is better than that without pruning. For some data sets, such as Amazon-Google and BeerAdvo-RateBeer, the F1 score has been significantly improved.

From the experiment result, we can see that the proposed pruning method is effective. Although the result of entity matching using the LWCR method is poor, this method can give an approximate similarity for all record pairs. If the approximate similarity of a record pair is too low, it means that this record pair has a high probability of being mismatched. Even if a small number of record pairs are matched with low similarity, they have a negative impact on constructing classifiers because all of their attributes have a low similarity. Therefore, using this pruning method can not only reduce the number of mismatched samples but also remove some matched pairs with low similarity, which prevents the query strategies from selecting "noise samples" to the labeled pool and improve the generalization ability of classifiers.

## 5. Concluding Remarks

In this paper we propose an active learning method with an integrated query strategy, i.e., ALMatcher, which can be well applied to entity matching tasks. This method can construct a machine learning model through only a small number of labeled samples, and achieve a high F1 score. In small-scale data sets, the effect of ALMatcher is even better than the DL-based methods. ALMatcher can effectively reduce the workload of labeling samples, and select the most valuable samples to experts for labeling. We propose a hybrid uncertainty as query strategy which is the suitable for entity matching tasks. The proposed method of initializing the labeled pool also has a good effect, which makes the classifiers fit fast. In addition, we attempt to use the method of linear weighted combination rules (LWCR) to prune the mismatched record pairs on the training set. The experiment shows the performance of the proposed method has been significantly improved after pruning.

ALMatcher can be well applied to entity matching tasks in new fields or lacking labeled samples. The ML based entity matching methods rely on a large number of labeled samples for training. The DL based methods usually need a pre-trained language model or domain-related data for fine-tuning. ALMatcher does not rely on domain knowledge or language model, and just uses a few labeled samples to get well performance. ALMatcher can greatly reduce the workload of labeling, and solve the limitations of traditional learning methods in entity matching tasks, which makes the new entity matching tasks lacking domain knowledge have an efficient and relatively accurate solution.

There are still some limitations on our work. Our stop criterion relies on the validation set, and the iteration is sometimes stopped prematurely in large-scale data sets. Besides, the proposed method only uses the structural similarity method for comparing record pairs, which does not consider the semantic similarity of attributes. In the future, we will further optimize the method of attribute similarity comparison, stop criterion, and pruning strategy. We will also further explore appropriate query strategies to make the active learning method more suitable for entity matching tasks.